# Rapid FinFET Modelling Using an Autoencoder

Amit Sarkar [1] Suman Sau[2] and Swagata Mandal[3]

[1] Indian Institute of Technology Delhi, Delhi, India
[2] ITER, siksha O anusandhan University bhubaneswar, Odisha, India
[3] Jalpaiguri Government Engineering College, W.B India
sarkaramit424@gmail.com

**Abstract.** This work presents a machine learning framework that leverages an autoencoder (AE) for the efficient modeling of FinFET. We first calibrated a BSIM-CMG model to generate a dataset of current-voltage ($I_D$-$V_G$) characteristics. This data was used to train an autoencoder that compresses full I-V curves into a low-dimensional latent space, which intrinsically encodes key device physics. A key innovation is the explicit incorporation of parameter such as drain to source voltage ($V_{DS}$) as an input feature, enhancing the model ability to capture bias dependent variation. The trained model successfully reconstructs full I-V curves and directly extracts critical device metrics including threshold voltage ($V_{TH}$), subthreshold slope (*SS*), and peak transconductance ($g_m$). This approach demonstrates that data driven compact models, built from actual characterization data, can achieve high accuracy with minimal training data, providing a powerful tool for rapid device characterization, modelling and circuit level simulation.

**Keywords:** FinFET, Autoencoder (AE), AI, Machine Learning

## 1 INTRODUCTION

The scaling of semiconductor devices, particularly with FinFETs and nanowires, has introduced unprecedented complexity in terms of understanding the physics and modelling of these devices [1-2]. Technology Computer-Aided Design (TCAD) simulations are indispensable for predicting device behaviour but are often limited by significant computational costs, convergence difficulties, and the need for extensive domain expertise to set up and interpret physical models. In recent years, machine learning (ML) has emerged as a powerful tool to augment TCAD, offering pathways to accelerate simulations, enable inverse design, and extract physical insights from data with minimal human intervention. However, a central challenge remains, how to develop models that are not only accurate within their training domain but can also generalize beyond it, all while minimizing the reliance on specialized domain knowledge. One prominent approach involves using neural networks to learn directly from TCAD-generated data, such as current-voltage (I-V) and capacitance-voltage (C-V) curves. For instance, it was demonstrated that an autoencoder (AE) could effectively compress full I-V and C-V curves of FinFETs into a low-dimensional latent space, from which device parameters could be accurately regressed using simple polynomial models [2]. This approach required as few as 25-50 training samples and removed the need for manual feature extraction (e.g., threshold voltage, subthreshold slope), thus significantly reducing the dependency on domain expertise. The latent variables were found to correlate strongly with physical parameters like gate length, fin width, and work function, enabling the prediction of not only full curves but also derived metrics such as transconductance ($g_m$) and drain-induced barrier lowering (DIBL).
Another innovative direction is the use of Physics-Informed Neural Networks (PINNs), which incorporate physical laws directly into the learning process [3]. Lu et al. showcased a PINN that could predict electrostatic potential and electron density profiles in a silicon nanowire transistor far beyond its training range (e.g., predicting inversion region behaviour from subthreshold data). By embedding physical constraints such as Poisson's equation and Fermi-Dirac statistics into the loss function, the model learned the underlying device physics without explicit access to TCAD solvers. This method demonstrated robust out of training range prediction capabilities, generalizing to gate voltages up to 10 times larger than those seen during training.

Further studies have contrasted ML approaches with and without domain expertise. Dhillon et al. applied both strategies to experimental gallium oxide Schottky barrier diodes, showing that while with domain expertise methods performed well, without domain expertise methods using denoising autoencoders could achieve comparable accuracy without manual feature engineering [4]. This is particularly valuable for novel devices where underlying physics may not be fully understood. Similarly, Wong et al. developed a TCAD-ML framework using principal component analysis (PCA) and polynomial regression to analyse device variations and operating temperatures from experimental I-V curves, showing the importance of dimensionality reduction in handling data [5].

Building on these advancements, our work implements an autoencoder based framework for predicting FinFET I-V characteristics and extracting key device parameters. The work leverages the principles of manifold learning, as advocated in the cited works, to compress I-V curves into a latent representation. Notably, it extends typical approaches by incorporating parameter such as drain to source voltage ($V_{DS}$) as an explicit input feature, enhancing the model's ability to capture various parameter like bias dependent effects. The autoencoder is trained on calibrated BSIM-CMG model generated data, and its performance is evaluated through the reconstruction of I-V curves and the extraction of critical metrics like threshold voltage ($V_{TH}$), subthreshold slope (*SS*), and peak transconductance ($g_m$). This approach provides a tool for rapid device modelling directly applicable to circuit level applications.

## 2 DEVICE SPECIFICATIONS AND BSIM-CMG CALIBRATION

The calibration of the BSIM-CMG model for the 12-nm p-channel bulk FinFETs was carried out following the parameter extraction guidelines provided in the BSIM-CMG 110.0.0 Technical Manual [6]. The devices under study featured a designed gate length of 14 nm, fin height of 30.5 nm, fin width of 14 nm, two fins, and an effective total width of 150 nm, as reported in the study [1]. The device specification are given in Table 1.

**Table 1. Device Specifications**

| Parameter | Value |
|---|---|
| Device Type | p-type |
| Gate Length ($\boldsymbol{L_G}$) | 14 nm |
| Fin Height ($\boldsymbol{H_{FIN}}$) | 30.5 nm |
| Fin Width ($\boldsymbol{W_{FIN}}$) | 14 nm |
| Number of Fins ($\boldsymbol{N_{FIN}}$) | 2 |
| Number of Gate Fingers ($\boldsymbol{N_{FINGER}}$) | 1 |
| Total Width ($\boldsymbol{W_{TOTAL}}$) | 150 nm |

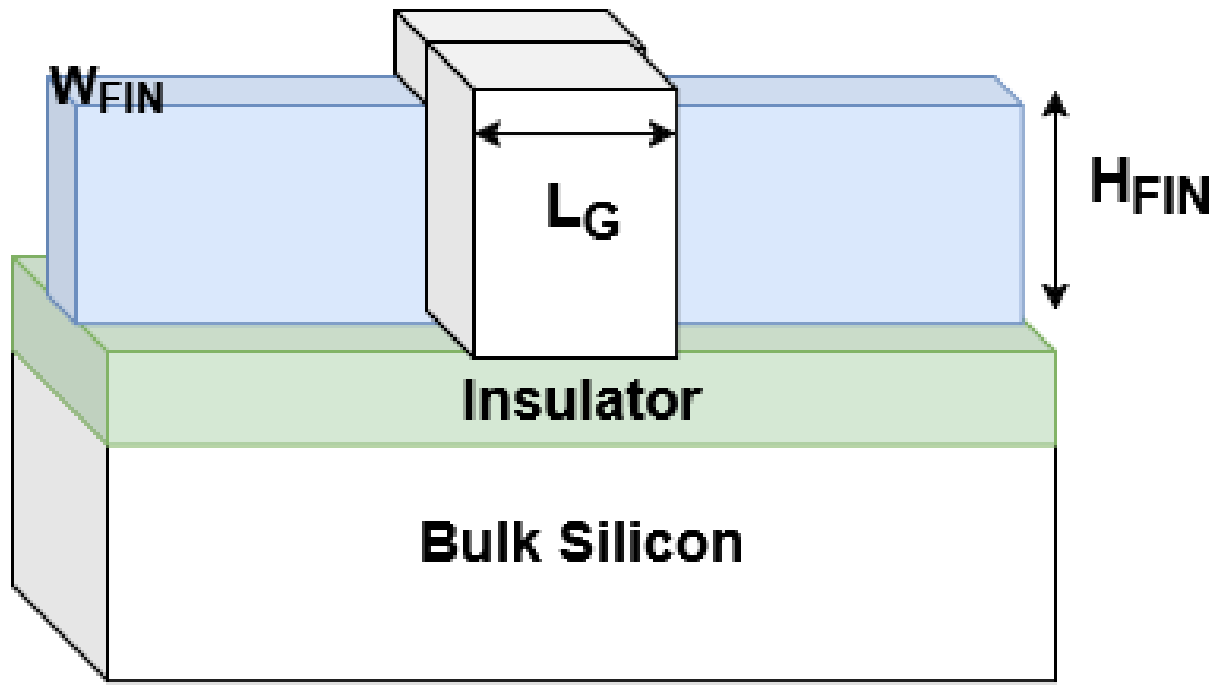

Figure 1: FinFET

## 3 MEATHODLOGY FOR MODELLING

The ($I_D$-$V_G$) dataset for this work was generated using a calibrated BSIM-CMG model, an industry-standard compact model whose parameters were tuned to correlate closely with experimental data from advanced FinFET devices [1]. This calibrated model was then employed to generate drain current versus gate voltage ($I_D$-$V_G$) data, each sweep performed at a specific drain-to-source voltage ($V_{DS}$), with $V_G$ swept from 0 V to -0.8 V, and the resultant data stored in individual files. The subsequent data preprocessing was performed for preparing raw data for neural network. To establish a consistent input structure, all curves were interpolated onto a common, finely-spaced $V_G$ vector, with extrapolated values clipped to a minimum current of $10^{-20}$ A. Given the drain current's span across several orders of magnitude, a base-10 logarithmic transformation was applied to compress the dynamic range, significantly enhancing training. The preprocessed dataset was then partitioned into a training pool and a test set, and the training data was used to compute the mean (μ) and standard deviation (σ) for both the log-scaled $I_D$ curves and the $V_{DS}$ values, enabling z-score normalization of the entire dataset to ensure all features were on a common scale.
The network architecture features a symmetric encoder-decoder structure: the input layer accepts a flattened vector of number of $V_G$ points + 1, representing a normalized $I_D$ curve concatenated with its corresponding normalized $V_D$ value. The encoder comprises two fully connected hidden layers, each containing 50 neurons with Rectified Linear Unit (ReLU) activation functions, which progressively compress the input into a compact latent space bottleneck of only three neurons, capturing the essential underlying factors of variation; the decoder mirrors this structure, utilizing two fully connected layers to reconstruct the original input vector from this compressed representation. This model was trained using the Adam optimizer for a maximum of 500 epochs, with a iterative strategy to identify the optimal model.

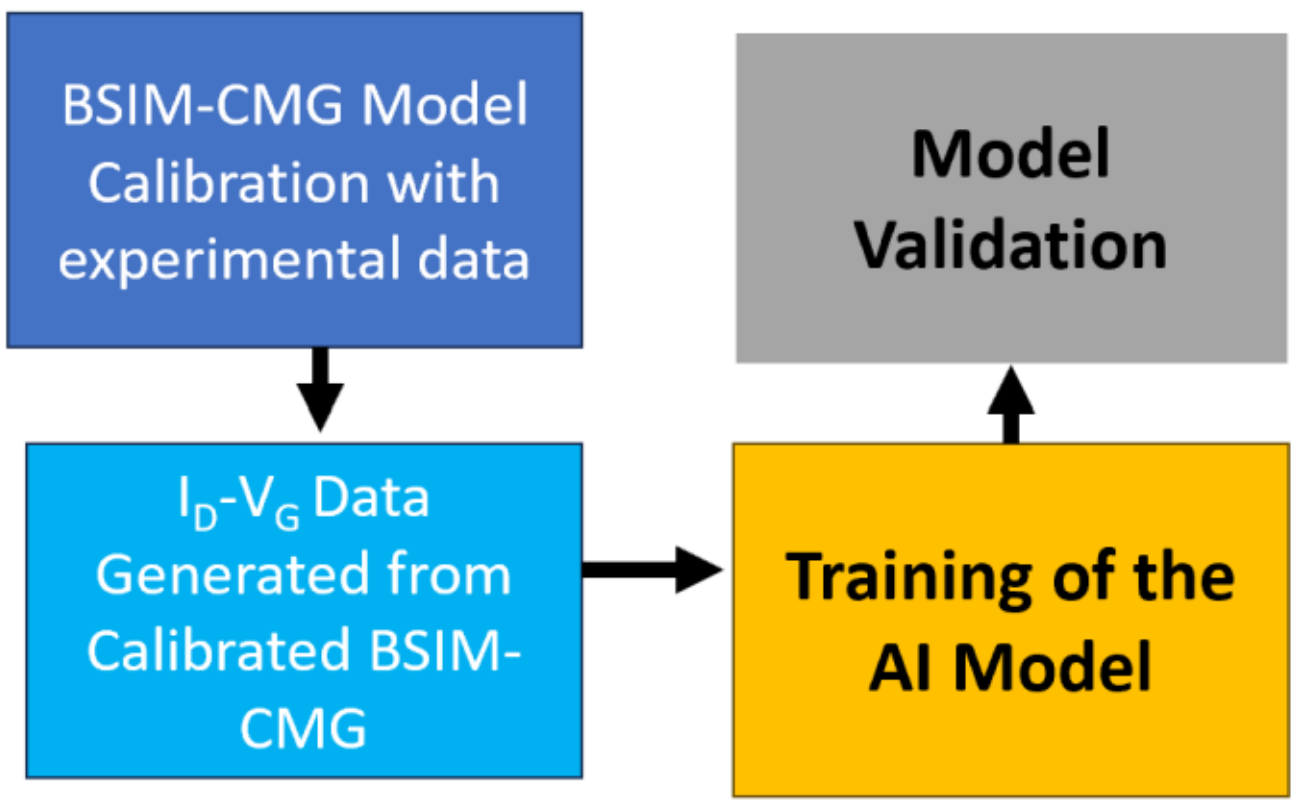

Figure 2: Methodology

The evaluation of the autoencoder extended far beyond mere reconstruction error, investigating its ability to accurately reproduce key transistor parameter extracted from the reconstructed $I_D$-$V_G$ curves. These metrics included the threshold voltage ($V_{TH}$), extracted using a constant current method, the subthreshold swing (*SS*) and the peak transconductance ($g_{m,max}$). The percentage error in these extracted parameters between the original BSIM-CMG curves and the autoencoder's reconstructions served as a metric for assessing the model accuracy and predictive capability, providing a validation of its performance. Figures 3 and 4 present the correlation between BSIM-CMG model and experimental data, and the model was calibrated with the experimental results to produce the dataset used for training the machine learning model.

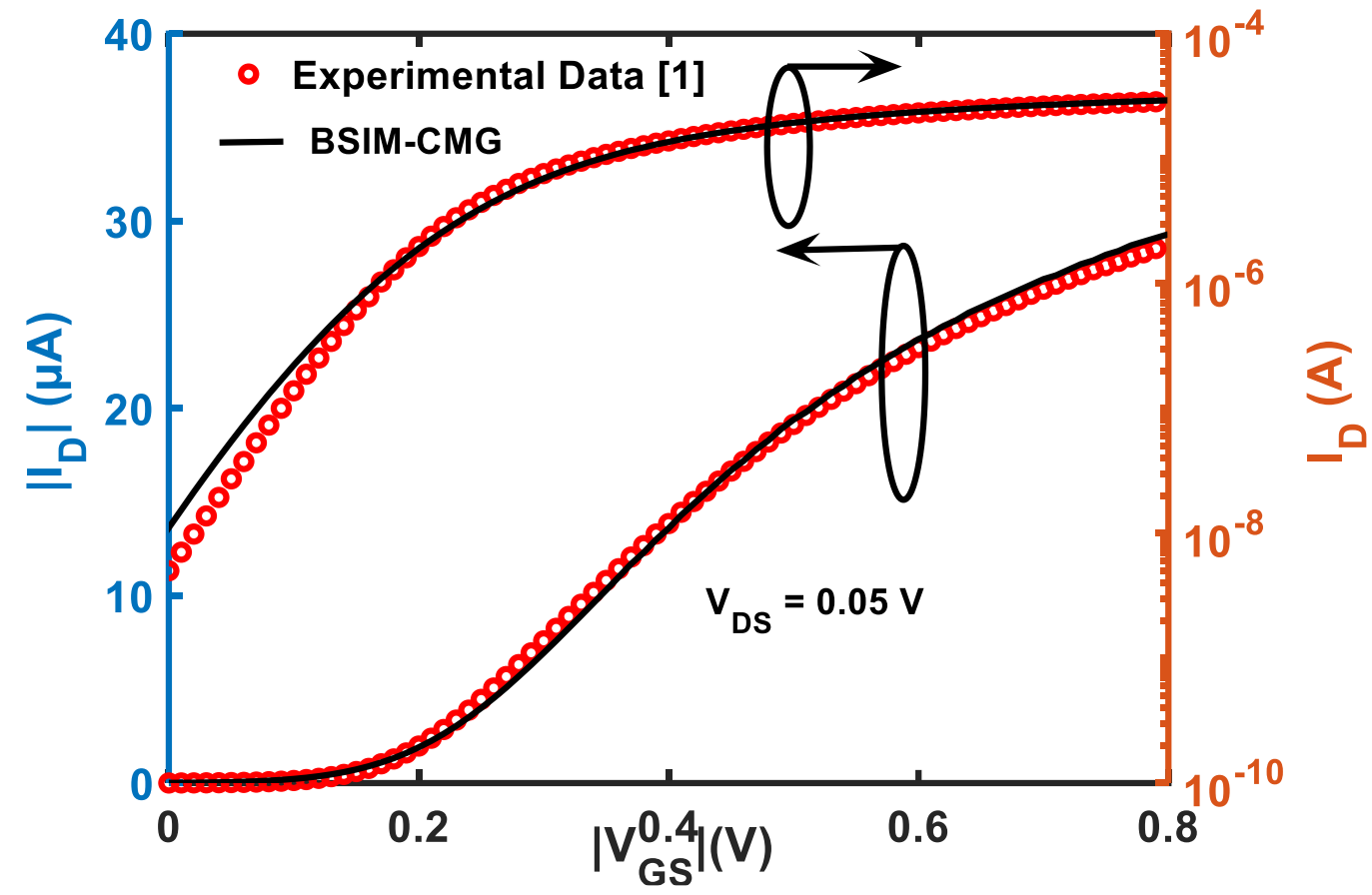

Figure 3: BSIM-CMG Model Hardware Correlation ($I_D$-$V_G$)

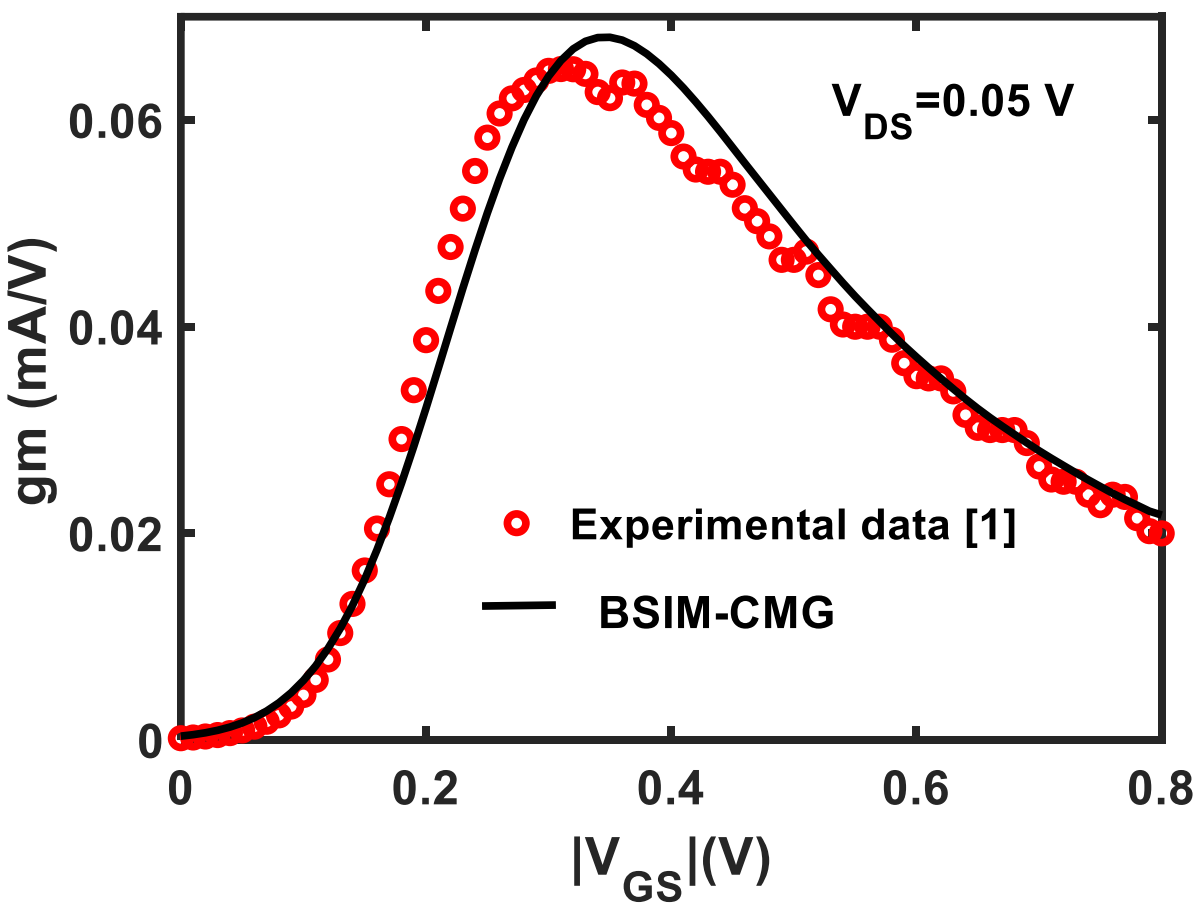

Figure 4: BSIM-CMG Model Hardware Correlation ($g_m$-$V_G$)

## 4 RESULT AND DISCUSSION

Figure 5 and 6 demonstrate a good correlation to validate AI-based model against the industry-standard BSIM-CMG model. As shown in Figure 5, the AI model's predictions for both drain current and transconductance are correlating well with BSIM-CMG across a range of gate voltages for $V_{DS}$ = 0.4 V taken for model validation. This exceptional accuracy is quantitatively validated in Figure 6, where the $R^2$ value consistently exceeds 0.9 across various drain to source voltages ($V_{DS}$), confirming the model accuracy. The $\Delta V_{TH}$ vs $V_{DS}$ plot illustrates the percentage error between the threshold voltage ($V_{TH}$) extracted from the original BSIM-CMG data and the $V_{TH}$ predicted by the AI model for multiple $V_{DS}$ value. A consistently low percentage error demonstrates that the AI model accurately captures the dependence of the threshold voltage on the drain bias. Similarly, the $\Delta SS$ vs $V_{DS}$ plot shows the percentage error in the predicted subthreshold swing.

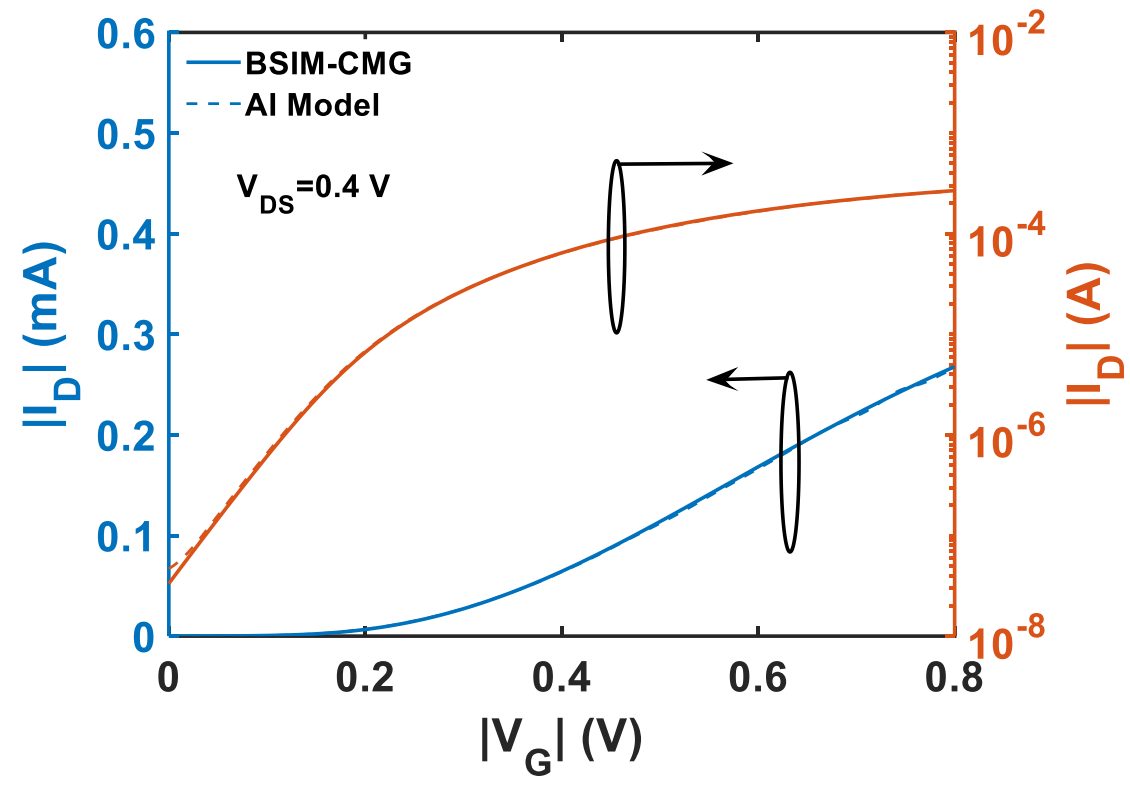

Figure 5: AI Model & BSIM CMG Correlation $I_D$-$V_G$

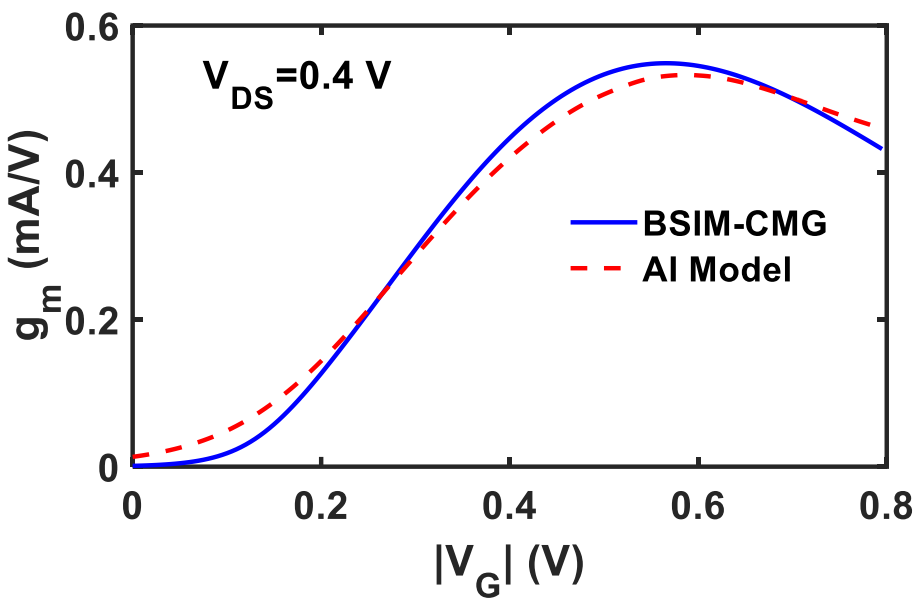

Figure 6: AI Model & BSIM CMG Correlation $g_m$-$V_{GS}$

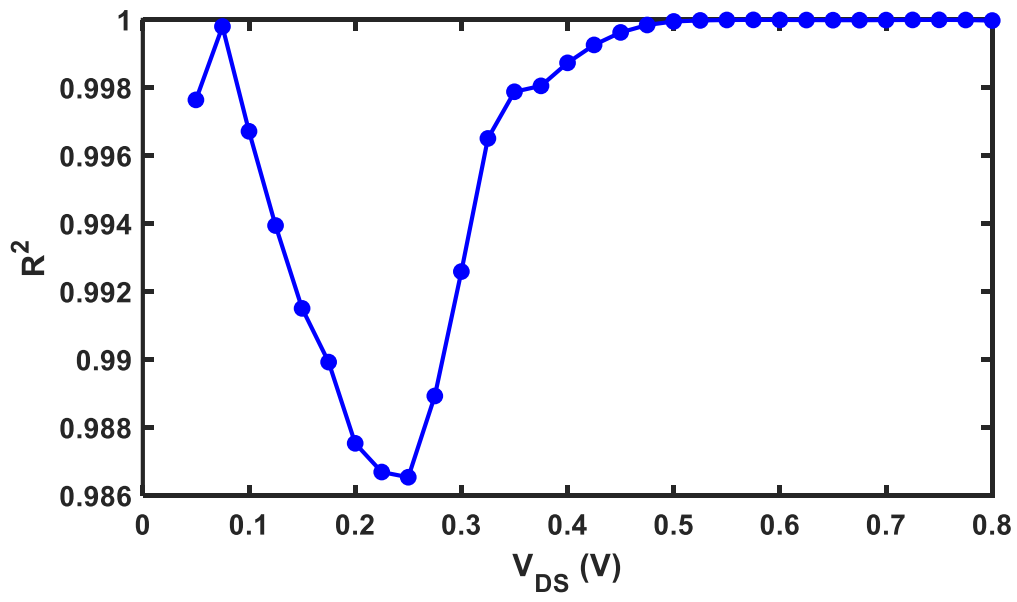

Figure 7: $R^2$ VS $V_{DS}$

A low error percentage was found in this plot and it is important as it confirms the model ability to precisely replicate the transistor's switching behaviour. The $\Delta g_m$ vs $V_{DS}$ plot demonstrates that the error in the AI model's prediction of peak transconductance remains consistently low, within a 0% to 5% range across the drain voltage sweep from 0 to 0.2 V and 0.4 to 0.8 V. While this error shows slightly more variability compared to $\Delta V_{TH}$ and ΔSS plot, expected since $g_m$ is a differential parameter highly sensitive to minor curve shape changes. This confirms the model accuracy and robustness, which is critical for Analog, digital and RF circuit design.

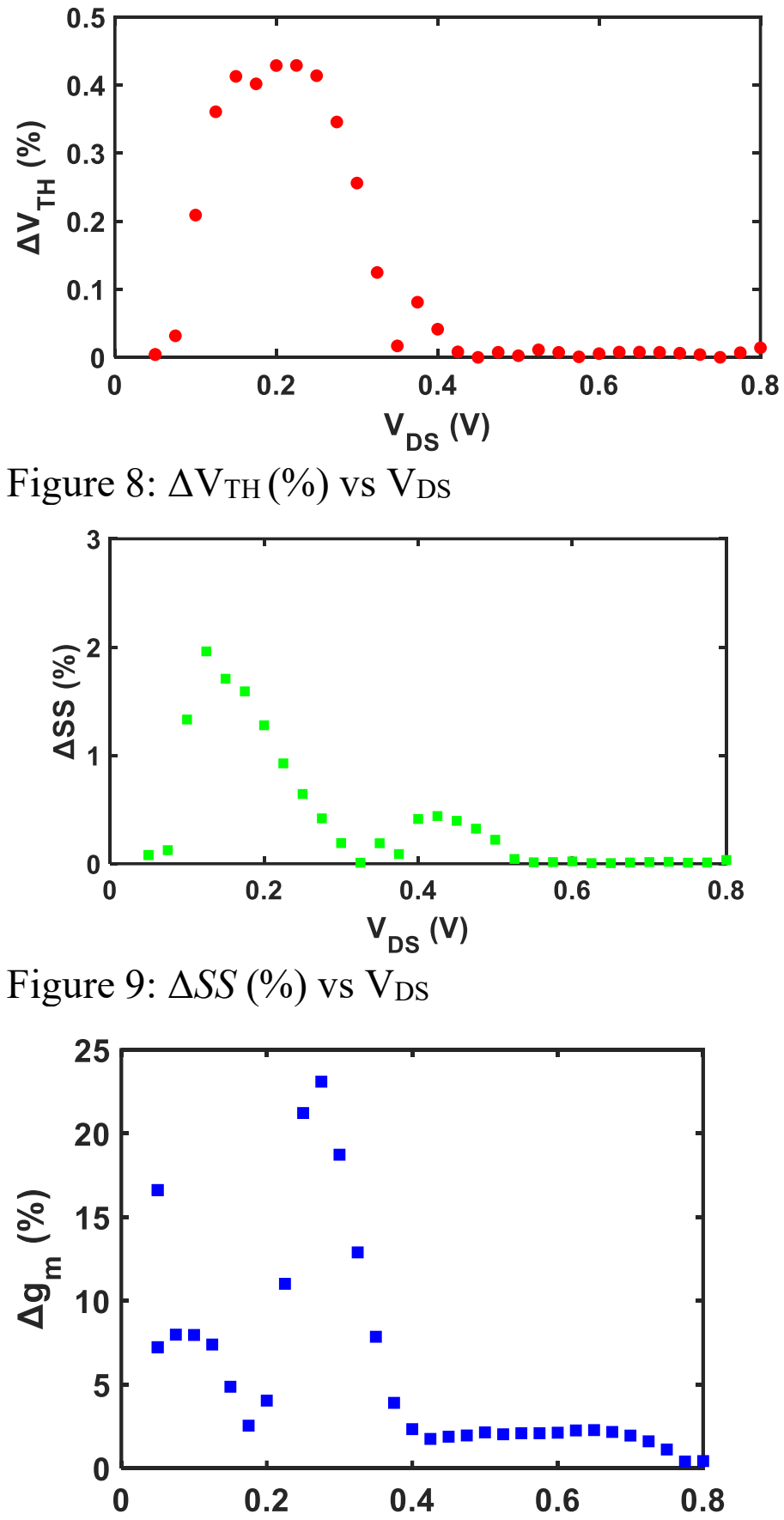

Figure 8: $\Delta V_{TH}$ (%) vs $V_{DS}$

Figure 9: Δ*SS* (%) vs $V_{DS}$

Figure 10: $\Delta g_m$(%) vs $V_{DS}$

# 5 CONCLUSION

The model validation results demonstrate that the proposed AI-based autoencoder model can accurately reproduce key device characteristics such as drain current, threshold voltage, subthreshold swing, and transconductance with consistently low prediction error. By accurately capturing characteristics of the device, the model ensures robustness of this approach, and its suitability for use in Analog, digital, and RF circuit design. This positions the autoencoder model as a alternative to compact models, offering faster modelling and accuracy for existing and upcoming device technology